\documentclass[10pt,conference,letterpaper]{IEEEtran}

\usepackage{amsmath,amsthm,amssymb}
\usepackage[ampersand]{easylist}
\usepackage{url}
\usepackage{array}
\ListProperties(Hide=100, Hang=true, Progressive=0ex, Style*=-- , Style2*=$\bullet$ ,Style3*=$\circ$ ,Style4*=\tiny$\blacksquare$)
\usepackage[textsize=footnotesize]{todonotes}
\usepackage{balance}

\newtheorem{thm}{Theorem}

\newtheorem{pro}{Problem}

\newtheorem{remark}{Remark}

\newcommand{\ptitle}[1]{\vspace{1mm}\noindent{\bf #1.}}
\newcommand{\ptitlenoskip}[1]{\noindent{\bf #1.}}

\newcommand{\bc}{\mathbf{c}}

\newcommand{\bw}{\mathbf{w}}
\newcommand{\bx}{\mathbf{x}}
\newcommand{\by}{\mathbf{y}}
\newcommand{\bI}{\mathbf{I}}

\newcommand{\bY}{\mathbf{Y}}

\newcommand{\dx}{{d_\bx}}
\newcommand{\dy}{{d_\by}}

\newcommand{\hv}{\hat{v}}

\newcommand{\blambda}{\boldsymbol{\lambda}}
\newcommand{\bmu}{\boldsymbol{\mu}}
\newcommand{\bSigma}{\boldsymbol{\Sigma}}

\newcommand{\bZero}{\mathbf{0}}

\newcommand{\hbx}{\hat{\bx}}

\newcommand{\hby}{\hat{\by}}
\newcommand{\hbY}{\hat{\bY}}

\newcommand{\bbR}{\mathbb{R}}

\newcommand{\cI}{\mathcal{I}}
\newcommand{\cC}{\mathcal{C}}

\newcommand{\cN}{\mathcal{N}}
\newcommand{\cX}{\mathcal{X}}

\title{Subjectively Interesting Subgroup Discovery\\on Real-valued Targets}

\author{%
{Jefrey Lijffijt{\small $^{*}$}, Bo Kang{\small $^{*}$}, Wouter Duivesteijn{\small $^{+}$}, Kai Puolam\"{a}ki{\small $^{=\#}$}, Emilia Oikarinen{\small $^{\#}$}, Tijl De Bie}{\small $^{*}$}%
\vspace{1.6mm}\\
\fontsize{10}{10}\selectfont\rmfamily\itshape
$^{*}$\,Department of Electronics and Information Systems, IDLab, Ghent University\\
Ghent, Belgium\\
\fontsize{9}{9}\selectfont\ttfamily\upshape
\,jefrey.lijffijt@ugent.be, bo.kang@ugent.be, tijl.debie@ugent.be%
\vspace{1.2mm}\\
\fontsize{10}{10}\selectfont\rmfamily\itshape
$^{+}$\,Department of Computer Science and Mathematics, Eindhoven University of Technology\\
Eindhoven, The Netherlands\\
\fontsize{9}{9}\selectfont\ttfamily\upshape
\,w.duivesteijn@tue.nl
\vspace{1.2mm}\\
\fontsize{10}{10}\selectfont\rmfamily\itshape
$^{=}$\,Department of Computer Science, Aalto University, Helsinki, Finland\\
\fontsize{9}{9}\selectfont\ttfamily\upshape
\, kai.puolamaki@aalto.fi
\vspace{1.2mm}\\
\fontsize{10}{10}\selectfont\rmfamily\itshape
$^{\#}$\,Finnish Institute of Occupational Health,
Helsinki, Finland\\
\fontsize{9}{9}\selectfont\ttfamily\upshape
\, emilia.oikarinen@ttl.fi
}

\begin{document}

\maketitle

\begin{abstract} \small\baselineskip=9pt Deriving insights from high-dimensional data is one of the core problems in data mining. The difficulty mainly stems from the fact that there are exponentially many variable combinations to potentially consider, and there are infinitely many if we consider weighted combinations, even for linear combinations. Hence, an obvious question is whether we can automate the search for interesting patterns and visualizations. In this paper, we consider the setting where a user wants to \emph{learn} as efficiently as possible about real-valued attributes. For example, to understand the distribution of crime rates in different geographic areas in terms of other (numerical, ordinal and/or categorical) variables that describe the areas. We introduce a method to find subgroups in the data that are maximally informative (in the formal Information Theoretic sense) with respect to a single or set of real-valued target attributes. The subgroup descriptions are in terms of a succinct set of arbitrarily-typed other attributes. The approach is based on the Subjective Interestingness framework FORSIED to enable the use of prior knowledge when finding most informative non-redundant patterns, and hence the method also supports iterative data mining.\end{abstract}

\section{Introduction}\label{sec:intro}

We introduce the central ideas by means of an example. Consider the situation that a user want to learn about crime demographics, based on the UCI Communities and Crime data\footnote{\url{http://archive.ics.uci.edu/ml/datasets/communities+and+crime}} \cite{redmond2002}. This data contains violent crime rates for all ($n = 1994$) districts in the US and over 120 other attributes describing demographic statistics of those districts. One method to learn about the relation between the `number of violent crimes' attribute and the demographic attributes is to extract \emph{subgroup patterns}, which are sets of data points where violent crime is surprisingly high (or low) and that share similar statistics for one or several demographic attributes. A subgroup pattern should be interpreted as `for data points that fall within the specified statistics that describe the subgroup, violent crime is surprisingly low/high'.

\begin{figure}[tp]
  \centering
  \includegraphics[width=0.9\columnwidth]{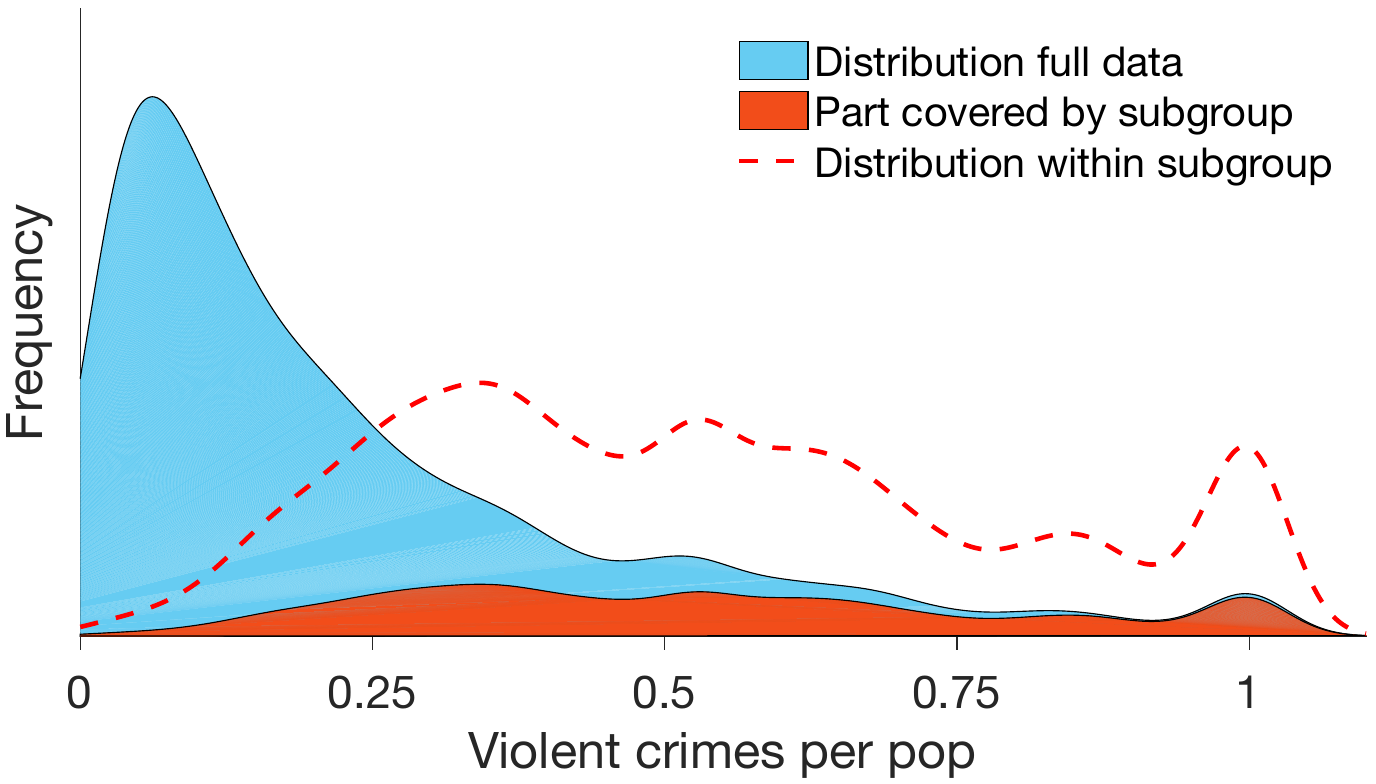}
  \caption{Distribution of violent crime over the full data (light blue area), part covered by the subgroup `high rate of unmarried mothers' (red area), and distribution within the subgroup (red dotted line). Height of  colored areas given by Gaussian-kernel smoothed estimates. The subgroup clearly covers a substantial amount of the data where the violent crime rate is relatively high.\label{fig:example-subgroup}}
\end{figure}

For example, the top subgroup pattern---identified through the method introduced in this paper---states that there are high violent crime rates in districts where many mothers are unmarried at the moment they give birth to their child (condition $PctIlleg >= 0.39$; mean violent crime rate $0.53$ in subgroup vs. $0.24$ overall). An illustration of the data coverage for this pattern is given in Fig.~\ref{fig:example-subgroup}. The subgroup covers $20.5\%$ of the data and may be interesting because the distribution of crime rates within this subgroup deviates substantially from the full data. If a user would have no prior expectations about the data, this pattern is highly informative.

Indeed, we may quantify how informative/interesting it is, in the Information Theoretic sense: the number of bits of information we gain about the data by learning about this pattern, which depends on the amount of data covered (more is better) and how much the distribution in the subgroup differs from our expectation (more is better; in this paper we consider mean and variance statistics). Typically, we would like to weight this against how complex the description of the pattern is (number of attributes used to describe the subgroup plus the number of statistics presented to the user, fewer is better), such that our aim is to provide a maximal \emph{information rate}.

This is precisely the contribution of this paper. We quantify the Information Content (IC; the amount of information gained) and Description Length (DL; the complexity of the description) for \emph{subgroup patterns}. However, while the example above has only one \emph{target attribute} (the violent crime rate), we also do this for multivariate real-valued targets, in order to enable users to learn about multivariate distributions. Besides, while the example above is about a surprisingly high mean (violent crime rate), we quantify the IC and DL for both mean and (co-)variance statistics.

As hinted at in the example, the IC of a pattern is inherently \emph{subjective}, i.e., particular to a user, because \emph{how much you learn depends on your prior knowledge}. We implement this subjectivity by modeling a background distribution over the data space that is a Maximum Entropy distribution subject to constraints corresponding to the current knowledge of a user. This approach is known as FORSIED \cite{debie2011,debie2013} and also immediately enables iterative mining of non-redundant patterns without much additional effort.

We have implemented an algorithm to iteratively mine interesting patterns which is freely available as open source code. We have not studied the algorithmic problem in detail, but the implementation is based on beam search, a frequently employed approach in subgroup discovery. That is, it maintains a list of most interesting patterns of arity $k$, expands these to arity $k+1$ and selects the most interesting patterns again. Ultimately, it outputs the most interesting pattern found. It handles categorical, ordinal, and numerical \emph{description attributes} (the demographic attributes in the example) and supports time constraints (e.g., stop after 1 minute of mining). The implementation is based on Cortana \cite{2011Meeng}.

In summary, this paper contributes the following:
\begin{easylist}
& We define a new pattern syntax for subgroups with a multivariate real-valued target distribution, called \emph{location}  and \emph{spread} patterns. (Sec.\@ \ref{sec:syntax})
& We introduce a method to quantify their interestingness in a subjective manner. (Sec.\@ \ref{sec:interestingness})
& Before that, we study how to incorporate prior knowledge into the background model, including previously identified patterns to enable iterative mining. (Sec.\@ \ref{sec:background})
& We present how to mine high-quality patterns using beam search and gradient descent. (Sec.\@ \ref{sec:search})
& We provide empirical evidence on four datasets that we can effectively find interesting patterns. (Sec.\@ \ref{sec:experiments})
\end{easylist}

Discussion of related work is presented in Sec.~\ref{sec:relatedwork}, directions for future work and conclusions are given in Sec.~\ref{sec:discussion}. All code, including  code for repeating the experiments, and links to the datasets are available at:  \url{https://www.dropbox.com/sh/3m1cgt1mh15k8bu/AAAViZtu5aeSOA3ybCS5mi-ta?dl=0}.

\section{Methods}\label{sec:problem}

\ptitlenoskip{Overview}
The high-level problem addressed in this paper is:
\begin{pro}\label{pro:main}{\em Main Problem.}
  Iteratively inform the user about the mean and variance of subsets of data points
	that can be described concisely in terms of the description attributes,
	such that the rate of information gain of the user about the target attributes is
  maximized at each iteration.
\end{pro}
We first formalize the type of \emph{pattern} shown to the user (Sec.~\ref{sec:syntax}). To explain how to find the most interesting patterns of this type (Sec.~\ref{sec:search}), we first need to formalize the background distribution (Sec. \ref{sec:background}) and the interestingness of patterns (Sec.~\ref{sec:interestingness}).

The formalization follows the FORSIED approach: we formalize the user's belief state about the target attributes by means of a background distribution, and quantify the IC of a pattern as the information (in its formal sense) the user gains about the target attributes by seeing the pattern.
The Subjective Interestingness (SI) of a pattern is then formalized as the (subjective) IC divided by the DL of the pattern.

\ptitle{Notation}
The data consists of a set of $n$ pairs $(\hbx_i,\hby_i)$, $i\in[n]$ (where $[n]$ is shorthand for $\{1,2,\ldots,n\}$), called the \emph{data points}.
Here, the so-called \emph{description attributes} of the $i$th data point $\hbx_i\in\prod_{j=1:\dx}\cX_j$ is assumed to be a tuple of $\dx$ attributes with domains $\cX_j$,
and $\hby_i\in\bbR^{\dy}$ is a vector containing the values for $\dy$ real-valued \emph{target attributes}.
We denote $\hbY=\left(\hby_1',\hby_2',\cdots,\hby_n'\right)'$.
In our setup the user
is interested in gaining an understanding of the behavior of the target attributes
in terms of the descriptions.

For example, the target attributes could contain healthcare-related attributes,
whereas the description attributes could describe lifestyle choices (e.g., smoking or not, sedentary or active lifestyle, etc).
Then, our method would yield insights into the healthcare target attributes, in terms of the lifestyle descriptions. In the example in Section \ref{sec:intro}, there is one target attribute (the violent crime rate) and over 120 description attributes.

We use hatted symbols to indicate these are empirical values.
Non-hatted equivalents will be used to denote the respective random variables, e.g., ${\bf Y}$.
They allow us to reason about the amount of uncertainty the user has about the data points.
In general, standard face lower case symbols denote scalars,
bold face lower case symbols denote tuples or vectors,
upper case bold face symbols denote matrices,
and upper case calligraphic letters denote sets.

\begin{figure*}[tp]
  \includegraphics[width=\textwidth]{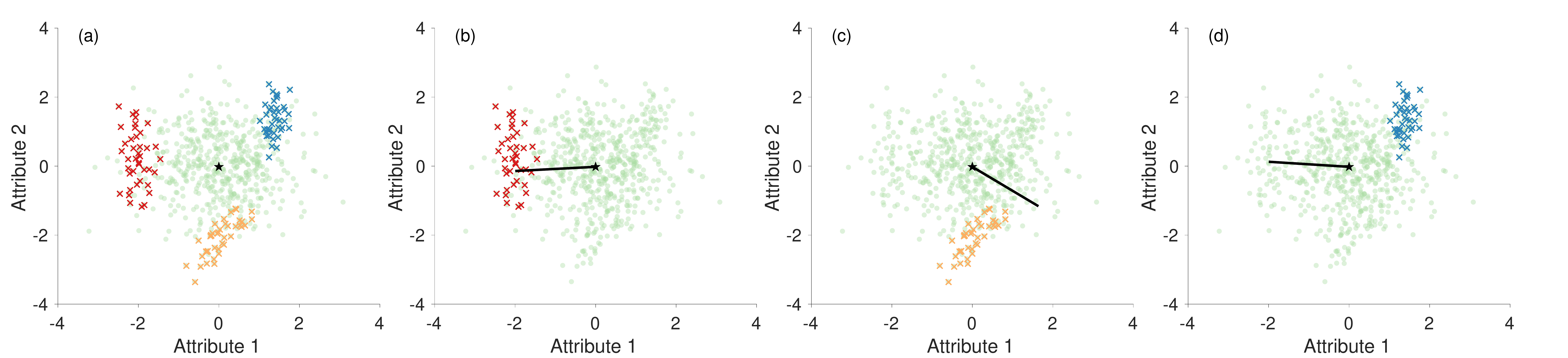}
  \caption{Patterns found in the synthetic data (\S \ref{sec:syntax},\S \ref{sec:syntheticData}), (a) Data with the embedded patterns highlighted. (b)---(d) Top ranked pattern discovered in iterations 1---3. Light green circles are random data points, darker colored crosses the three embedded clusters. The black star represents the mean of the full data and the black lines are the angles of the most surprising variance direction. The two axes correspond to the only two target attributes.
\label{fig:syntheticDataFigs}}
\end{figure*}

\subsection{Location and spread patterns\label{sec:syntax}}

\ptitlenoskip{Subgroups, intentions, and extensions} We define patterns in terms of \emph{subgroups}. A subgroup is defined by a set of \emph{conditions} on the description attributes (the value combination is the subgroup \emph{intention}) and by the set of data points for which the description attributes satisfy these conditions (the index set is the subgroup \emph{extension}).

The intention is described in a pre-defined formal \emph{description language}, such as in the form of a conjunction of conditions on individual metadata attributes. For $\cX_j=\bbR$, such conditions are typically inequality conditions, and for $\cX_j$ categorical they can be set in-/exclusion conditions. The extension is then specified by the index set $\cI\subseteq[n]$ with $i\in \cI$ iff $\hbx_i$ satisfies the conditions.

\ptitle{Location and spread patterns}
Subgroups tend to be informative if the target attribute values of data points in the extension $\{\hby_i|i\in \cI\}$ are \emph{unusual} in some sense.
The way in which this set is unusual will be quantified by means of statistics---functions of this set of data points.
For example, its empirical mean could be unusually far from what the user would expect,
or its empirical variance around this mean could be unusually small or large along a certain direction.

To be precise, let us define two statistics
$f_\cI:\bbR^{n\times\dy}\mapsto\bbR^{\dy}$ and
$g^\bw_\cI:\bbR^{n\times\dy}\mapsto\bbR$ as follows:
\begin{eqnarray}
f_\cI(\bY)&=&\sum\nolimits_{i\in \cI}{\by_i}/\left|\cI\right|,\mbox{ and}\label{eq:mean} \\
g^\bw_\cI(\bY)&=&\sum\nolimits_{i\in \cI}
{\left(\left(\by_i-\hat\by_\cI\right)'\bw\right)^2}/\left|\cI\right|, \label{eq:var}
\end{eqnarray}
where $\hat\by_\cI=\sum_{i\in \cI}{\hat\by_i}/\left|\cI\right|$ and ${\bf
w}\in{\mathbb{R}}^k$ is a unit vector, i.e., ${\bf w}'{\bf w}=1$.
The first statistic (actually a set of $\dy$ statistics), when evaluated on $\hbY$, quantifies the average vector of the data points in the extension (i.e., its average \emph{location}), whereas the second quantifies the spread around that location.
Patterns considered here are specified by an intention, which uniquely determines the extension $\cI\subseteq[n]$,
a unit vector $\bw$, and the specification of the empirical values of one or both of the statistics $f_\cI(\hbY)$ and $g^\bw_\cI(\hbY)$:
we call it a \emph{location pattern} when the former is specified, and a \emph{spread pattern} when the latter is specified.
We find that the spread of a subgroup cannot be interpreted straightforwardly without knowing its location, hence
we only ever provide the user with spread patterns for subgroups for which the location pattern has been provided first. That is, we only explain the (co-)variance structure of subgroups for which the user already knows the precise mean value within the subgroup for all attributes. 

\ptitle{Example} For the synthetic data shown in
Fig~\ref{fig:syntheticDataFigs}a, a location pattern is an intention, e.g., `Attribute3 = true', along with
the mean of the subgroup, e.g., the dark red set of points. A spread pattern is an
intention, a direction (a weight vector of unit length, as in Fig.~\ref{fig:syntheticDataFigs}b), and the
magnitude of the variance in that direction.

\subsection{Modelling the user's belief state}\label{sec:background}

As we are interested in quantifying how informative a pattern is \emph{to a particular user},
we quantify its informativeness (the IC) with respect to a model for the user's belief state.
Patterns that contrast more strongly w.r.t.\@ this belief state are more surprising and thus carry more information for the user. We model the user's belief state by the means of a so-called \emph{background distribution},
represented by a density function $p$.
This is a distribution over the possible data values (here, a distribution for  $\bY$),
which assigns a higher probability density to data values that are deemed more probable by the user. The general form of this approach is known as FORSIED \cite{debie2011,debie2013}.

The initial background distribution, with density function $p_0$, can be
estimated as the distribution of Maximum Entropy (MaxEnt) subject to constraints that
express the user's knowledge, aka. the prior beliefs. The reason to use the MaxEnt distribution is that this is the only neutral choice, i.e., the only distribution that contains no other information \cite{Cover2005}. Importantly, during the mining process the background
distribution evolves, as each pattern shown to the user changes their
belief state about the data. We first derive the initial background
distribution, and then show how it can be updated to account for location and
spread patterns.

\ptitlenoskip{Initial background distribution}
To derive the initial background distribution, we need to assume what prior beliefs the user may have.
We consider the case where the user expects the overall mean of $\hbY$ to be equal to a specified vector $\bmu$,
and its covariance to be equal to a specified matrix $\bSigma$. Notice that these need not be equal to the empirical statistics; they may be anything.
The MaxEnt distribution subject to such expectations is well-known to equal a multivariate Normal distribution with $\bmu$ and $\bSigma$ as parameters:
\begin{align}
p_0(\bY)  \propto \exp{\left(-\sum\nolimits_{i=1}^n{
\left(\by_i-\bmu\right)'\bSigma^{-1}\left(\by_i-\bmu\right)
}/2
\right)}.
\end{align}

\ptitlenoskip{The evolving background distribution}
Given a pattern, the background distribution has to be updated to reflect the user's acquired knowledge.
This can be done by minimally altering the background distribution while ensuring the statistic is (in expectation) as specified by the pattern.
Here, \emph{minimally} is naturally measured in terms of the Kullback-Leibler (KL) divergence. This approach is known as the \emph{principle of minimum discrimination information}, a generalization of the MaxEnt principle.

We postulate, for now, that through subsequent updates in this way,
the background distribution will continue to be a product of multivariate Normal distributions,
although the means and covariances of the different data points may differ. I.e., after $t$ iterations, the density function of the background distribution will be:
\begin{align}\label{eq:expfam}
p_t(\bY) & \propto \exp{\left(-\sum\nolimits_{i=1}^n{
\left(\by_i-\bmu_i^t\right)'(\bSigma_i^t)^{-1}\left(\by_i-\bmu_i^t\right)
}/2
\right)},
\end{align}
where data points may have differing means $\bmu_i^t$ and covariance matrices $\bSigma_i^t$.
This holds for $t=0$ (when $\bmu_i^0=\bmu$ and $\bSigma_i^0=\bSigma$ for all $i$),
and the following shows that updating a distribution to account for location and spread patterns merely changes the parameter values,
leaving the distribution's parametric form intact.

\ptitle{Background distribution updating for location patterns}
To update $p_t$ given a location pattern for a subgroup with extension $\cI_{t+1}$,
we must solve the following optimization problem:
\begin{equation}
p_{t+1} = \arg\min_{q} KL(q\mid\mid p_t)=E_{q}\left[\log{\left(q(\bY)/p_t(\bY)\right)}\right]
\end{equation}
\begin{equation}\label{eq:meancons}
\text{subject to}\qquad E_{q}\left[f_{\cI_{t+1}}(\bY)\right]=\hby_{\cI_{t+1}},
\end{equation}
with the additional technical constraint $E_q\left[1\right]=1$ that guarantees that the
distribution has a proper normalization.

\begin{thm}
Let $p_t$ be a density function of the form of Eq.~\eqref{eq:expfam}.
Then, $p_{t+1}$ has the same parametric form, with:
\begin{align}
  \bmu_i^{t+1} &= \bmu_i^t+
  \sum\nolimits_{i\in \cI_{t+1}}(\hby_{\cI_{t+1}}-\bmu_i)/\left|\cI_{t+1}\right| ,
\end{align}
for $i\in \cI_{t+1}$, and all other parameters unaltered.
\end{thm}
\begin{IEEEproof}[Proof (outline only for brevity)]
Given the convexity of the KL-divergence and the linearity of the constraints, the optimization problem to be solved is convex and any stationary point is a global minimum.
The Karush-Kuhn-Tucker (KKT) stationarity condition gives us the functional form of $p_{t+1}$:
\begin{align}
p_{t+1} &\propto p_{t} \exp{\left(-\blambda'\sum\nolimits_{i\in \cI_{t+1}}\by_i\right)},
\end{align}
for a vector of KKT multipliers $\blambda$.
Manipulating this expression shows that $p_{t+1}$ is still of the form of Eq.~(\ref{eq:expfam}),
with $\bmu_i^{t+1}=\bmu_i^t+\bSigma_i^t\blambda$ for $i\in \cI_{t+1}$ and all other parameters unaltered.
The optimal value of $\blambda$ can be found by ensuring primal feasibility,
yielding that $\blambda=\sum_{i\in \cI_{t+1}}(\bSigma_i^{t})^{-1}(\hby_{\cI_{t+1}}-\bmu_i)$.
Substituting this for $\blambda$ in the expression for $\bmu_i^{t+1}$ proves the theorem.
\end{IEEEproof}

\ptitle{Background distribution updating for spread patterns}
To update the background distribution given a spread pattern for a subgroup with extension $\cI_{t+1}$,
we need to use the constraint
\begin{equation}\label{eq:varcons}
E_{q}\left[g^\bw_{\cI_{t+1}}(\bY)\right]=\hv^\bw_{\cI_{t+1}},
\end{equation}
in the KL-minimization problem, where for conciseness we denote the empirical variance as $\hv^\bw_{\cI_{t+1}}\triangleq g^\bw_{\cI_{t+1}}(\hbY)$.

\begin{thm}
Let $p_t$ be a density function of the form of Eq.~\eqref{eq:expfam}.
Then, $p_{t+1}$, updated for a spread pattern with spread $\hv^\bw_{\cI_{t+1}}$, has the same parametric form,
with:
\begin{align}
\bmu_i^{t+1} &= \bmu_i^t + \lambda\bw'(\hby_{\cI_{t+1}}-\bmu_i^t)\bSigma_i^t\bw/\left(1+\lambda\bw'\bSigma_i^t\bw\right),\\
\bSigma_i^{t+1} &= \bSigma_i^t - \lambda\bSigma_i^t\bw\bw'\bSigma_i^t/\left(1+\lambda\bw'\bSigma_i^t\bw\right),
\end{align}
for $i\in \cI_{t+1}$, and all other parameters unaltered.
The optimal value for $\lambda$ is found as the (unique) zero of the following equation:
\begin{align}
\begin{aligned}
\sum_{i\in \cI_{t+1}}\frac{\bw'\bSigma_i^t\bw}{1+\lambda\bw'\bSigma_i^t\bw} + \sum_{i\in \cI_{t+1}}&\left(\frac{\bw'(\hby-\mu_i^t)}{1+\lambda\bw'\bSigma_i^t\bw}\right)^2\\
&=|\cI_{t+1}|\hv^\bw_{\cI_{t+1}}.
\end{aligned}
\end{align}
\end{thm}
The proof is omitted for brevity. It is more tedious but analogous to the previous one.

\ptitle{Accounting for a set of location and spread patterns}
If we want to take into account a set of location and spread patterns,
the KL-divergence minimization problem needs to be solved with a constraint for each of these patterns.
The problem remains convex, however, such that a coordinate-descent approach converges to the global optimum.
This means iteratively updating the background distribution for each of the patterns, until convergence.
As long as the extensions of the different patterns have limited overlaps,
as is the case in our experiments, convergence occurs very rapidly.

\ptitle{Implementation details}
Rather than updating the parameters $\bmu_i$ and $\bSigma_i$,
we actually update the \emph{natural parameters} $\bSigma_i^{-1}\bmu_i$ and $-\frac{1}{2}\bSigma_i^{-1}$ of these multivariate Normal distributions.
This is numerically and computationally advantageous, but we feel it provides more insight to discuss the updates to $\bmu_i$ and $\bSigma_i$ above.

Also note that, maintaining and updating the background distribution may be costly if implemented naively. Each $\bmu_i^t$ and $\bSigma_i^t$ needs to be remembered and updating them involve summations over $\cI_{t+1}$ terms.
Yet, the number of \emph{distinct} $\bmu_i^t$ and $\bSigma_i^t$ remains limited.\footnote{Indeed, $\bmu_i^t=\bmu_j^t$ and $\bSigma_i^t=\bSigma_j^t$ for all $i$ and $j$ such $i,j\in \cI_s$ or $i,j\not\in \cI_s$ for all $s\in[t]$, since they will have been subjected to the same updates.}

\subsection{Subjective Interestingness}\label{sec:interestingness}

Given a background distribution, \cite{debie2011} proposed that the Subjective
Interestingness (SI) of a pattern can be computed as a ratio of two quantities:
(a) the \emph{Information Content} (IC) of a pattern, which is the negative log
probability of the pattern under the background distribution; and (b) the
\emph{Description Length} (DL), which measures the effort a user has to make to
understand and internalize the pattern.

To describe location patterns, we have to inform the user about the number of
conditions in the pattern's intention, the conditions themselves, and the mean
values for all attributes (to sufficient accuracy). For spread patterns, instead
of the means, the vector $\bw$ needs to be described, with its
magnitude. All these parts of the code have constant length, except for the set
of conditions, which has a length proportional to the number of conditions
$|\cC|$. Thus:
\begin{equation*}\label{eq:descrLength}
  \text{DescriptionLength} = \gamma |\cC| + \eta ~ (+1),
\end{equation*}
where the $+1$ applies to spread patterns only because they have one more term then location patterns.

We discuss determining $\gamma$ and $\eta$ in Remark~\ref{rem:parameters} below.
Note that it does not matter whether the DL is reflective of reality \emph{in absolute terms}, because
the actual SI scores are irrelevant. What matters is the ranking, hence
it is desirable that $\gamma$ is chosen well relative to $\eta$.

As the IC (thus the SI) depends on the pattern type, we derive it first
for location patterns and subsequently for spread patterns.

\ptitle{SI for location patterns} As the background distribution
\eqref{eq:expfam} for the target values $\bY$ of a data record is a normal
distribution, the marginal distribution $p_{f_\cI}$of the mean $f_\cI(\bY)$ of a
subgroup $\cI$ is again a normal distribution, with mean $\bmu_\cI = \sum_{i \in
\cI}\bmu_i/|\cI|,$ and covariance $\bSigma_\cI = \sum_{i\in
\cI}\bSigma_i/|\cI|$. The IC of a location pattern with extension $\cI$ is thus
the negative log probability of the pattern. Written in full:
\begin{align}
\begin{aligned}
IC_f(\cI) = &-\log{p_{f_\cI}\left(f_\cI(\bY)\right)} =  \log\left((2\pi)^\dy|\bSigma_\cI|\right)/2 \\
  &+ (f_\cI(\hbY) - \bmu_\cI)'\bSigma_\cI^{-1}(f_\cI(\hbY) - \bmu_\cI)/2.
\end{aligned}
\end{align}
The SI of a location pattern with extension $\cI$ and statistic $f_\cI$ reads:
\begin{equation}\label{eq:siLocation}
  SI_f(\cI) = IC_f(\cI)/\left(\gamma |\cC_\cI| + \eta\right).
\end{equation}

\ptitle{SI for spread patterns} While the SI of a location pattern can be
computed analytically, evaluating the SI for a spread pattern is more complex.
However, it can be approximated well.

If the patterns assimilated into the background so far do not overlap (i.e.,
non-intersecting extensions)\footnote{If the patterns used to update the
background distribution do overlap, then $\bmu_i \neq \hby_\cI$ even after the
update. So the random variable in Eq.~\eqref{eq:chiSquared} follows a
non-central chi-squared distribution, hence the linear combination
Eq.~\eqref{eq:linCombChiSquared} also changes. In this case, we approximate the
SI with the same computation for the non-overlapping situation. }, then after updating the background distribution
with location information of the pattern, the parameter $\bmu_i$ of the
background model equals the observed mean of subgroup $\hby_\cI$. So we can
derive:
\begin{align}
    (\by_i - \hby_\cI)'\bw(\bw'\bSigma_i\bw)^{-1/2} &\sim \cN(\bZero, \bI)\ \text{(normal distr.)}, \\
    ((\by_i - \hby_\cI)'\bw)^2/\left(\bw'\bSigma_i\bw\right) &\sim \chi^2_1\ \text{(Chi-squared, 1 d.f.)}.\label{eq:chiSquared}
\end{align}
Denote the chi-squared random variable derived above by $\bc_{i,1} = ((\by_i - \hby_\cI)'\bw)^2/(\bw'\bSigma_i\bw)$. Then, the variance statistic \eqref{eq:var} is a linear combination of chi-squared random variables:
\begin{equation}\label{eq:linCombChiSquared}
  g_\cI^{\bw}(\bY) =
  \sum\nolimits_{i\in \cI} \bw'\bSigma_i\bw \cdot \bc_{i,1}/|\cI|.
\end{equation}

The probability density function of a linear combination of chi-squared
distributed random variables has been studied extensively, but a closed form
analytic solution is unknown. Here we choose the state-of-art approximation
proposed by \cite{zhang2005approximate}: Writing $a_i$ for the coefficient
$\bw'\bSigma_i\bw/|\cI|$,
they prove that the distribution of $g_\cI^{\bw}(\bY)$ can be accurately
approximated by an affine function of a chi-squared random variable $\bc_m$ with
$m$ degrees of freedom:
\begin{equation}\label{eq:approxChisquared}
  g_\cI^{\bw}(\bY) = \alpha \bc_m + \beta,\ \ \text{where} \ \  \alpha = \frac{\sum_{i\in \cI_s}a_i^3}{\sum_{i\in \cI_s}a_i^2},
\end{equation}
\begin{equation*}
  \beta = \sum_{i\in \cI_s}a_i - \frac{\left(\sum_{i\in \cI_s}a_i^2\right)^2}{\sum_{i\in \cI_s}a_i^3},\ \
  m = \frac{\left(\sum_{i\in \cI_s}a_i^2\right)^3}{\left(\sum_{i\in \cI_s}a_i^3\right)^2}.
\end{equation*}
Therefore the approximated probability density function reads:
\begin{equation*}
  p_{g_\cI^{\bw}}(g_\cI^{\bw}(\bY)) \approx \frac{\left(\left(g_\cI^{\bw}(\bY)- \beta\right)/\alpha\right)^{\frac{m}{2}-1}e^{-\frac{g_\cI^{\bw}(\bY)-\beta}{2\alpha}}}{\left( \alpha\cdot2^{\frac{m}{2}}\Gamma(m/2) \right)}.
\end{equation*}
Thus the IC for a spread pattern with extension $\cI$ is given as:
\begin{align}\label{eq:icSpread}
  \text{IC}_g^\bw(\cI) = &-\log{p_{g_\cI^{\bw}}\left(g_\cI^{\bw}({\bf Y})\right)} \approx \log\left(2^{\frac{m}{2}}\Gamma(m/2)\right) \nonumber\\
  &+  \alpha -(m/2-1)\log\left(\left(g_\cI^{\bw}(\bY)  - \beta\right)/\alpha\right)  \nonumber\\
  &+ \left(g_\cI^\bw(\bY) -\beta\right)/\left(2\alpha\right).
\end{align}
The SI is then given by
\begin{equation}\label{eq:siSpread}
  SI_g^{\bf w}(\cI)= \text{IC}_g^\bw(\cI)/\left(\gamma |\cC_\cI| + \eta + 1\right).
\end{equation}

\begin{remark}\label{rem:parameters}
In practice, the SI's from Eqs.~(\ref{eq:siLocation}) and (\ref{eq:siSpread}) are only used for ranking the patterns,
or even just for finding the single most interesting pattern.
The absolute value of the SI is largely irrelevant in practice.
Thus, we can set $\eta=1$ without losing generality,
such that only $\gamma$ remains as a parameter, the value of which essentially depends on the `coding scheme' used to present the pattern to the user.

We do know of any principled approach to choose $\gamma$ well. Notice that the
problem here is not to do model selection in the statistical sense, but rather
the DL should be determined based on aspects of human cognition. In this paper,
we set $\gamma = 0.1$ throughout all the experiments. However, tuning $\gamma$
biases the results toward more or fewer conditions to describe the subgroup and
hence tuning could be useful.
\end{remark}

\subsection{Search strategies}\label{sec:search}
\ptitlenoskip{Overall approach}
We have not studied the complexity formally, but the optimization problem for
either pattern type appears to be very difficult. Tiling \cite{GGM:04}, a
similar and easier-appearing problem, is already NP-hard. The score function
here (the SI) is also not monotonic and, if the cardinality of metadata
attributes is large, pattern enumeration, which then equals exhaustive search,
is not a feasible strategy. For spread patterns, the search problem is
essentially a dimensionality reduction problem. From empirical results, we learn
that the search problem can have many local optima. Besides, there is no structure in
the problem that struck us as easy to use.

Hence, we resort to optimization procedures that are commonly used in either
scenario. In brief, to find location patterns that maximize
Eq.~\eqref{eq:siLocation}, we employ beam search. For spread patterns, we first
search for the best location pattern and after updating the background
distribution with the location, we use gradient descent to find the weight
vector $\bw$ that maximizes Eq.~\eqref{eq:siSpread} for that subgroup. The
procedures are outlined in more detail below.

\ptitle{Location pattern}
Beam search systematically explores the conjunctions of conditions by expanding
a limited set of conjunctions that have the largest SI so far. It evaluates
conjunctions of conditions on metadata attributes in a level-wise manner. On
each level, a limited list (beam width) of most promising combinations is
maintained. On the next level, the algorithm exhaustively grows the combinations
from the limited pattern list and maintains again the best. The mining process
stops when all possible conjunctions of conditions are explored or a chosen
stopping criterion is met, either a maximum search depth or time spent. Then,
the best pattern found throughout the search is given as output.

\ptitle{Spread pattern}
Finding the best spread pattern consists of two steps: (1) find the best
location pattern and update the background distribution with that information,
(2) for that location pattern find the most interesting direction in the target
space.
We have already described the first step; the second step can be formularized in
terms of the following optimization problem:
\begin{equation}\label{eq:maxsi}
  \max_{\bw}\ \ SI(g_{\cI}^{\bw}(\bY)) \qquad \qquad
  \text{s.t.}\ \ \bw'\bw = 1.
\end{equation}
Since the description length in $SI$ is fixed for a specific extension $\cI$,
the problem (\ref{eq:maxsi}) maximizes the entropy of a $\chi^2$ distribution
\eqref{eq:icSpread} over the unit sphere. To optimize $\bw$, we apply the
off-the-shelf manifold optimization tool Manopt \cite{2014Boumal} with the unit
sphere as the manifold, and solve it with the gradient-based solver.\footnote{We
computed the gradient analytically, but details are omitted due to lack of
space.}

\section{Experiments}\label{sec:experiments}

In this section we evaluate whether our method is able to find good location and spread patterns in terms of SI and whether the model updates work as expected.
We also studied the pattern descriptions, to see whether the patterns found appear to be interesting.  We conducted experiments on four datasets: one synthetic and three publicly available ones, of widely varying nature. The results for each dataset are described in the following subsections. The final subsection considers the scalability of the methods.

We used the beam search available within the data mining tool Cortana \cite{2011Meeng}, using
the following settings: descriptions on numerical metadata are based on $\geq$ and $\leq$ relations with four split points (1/5--4/5 percentiles). The beam width is set to 40 and the search depth is four conditions. The search logs the best 150 subgroups, with a maximum run time of 5 minutes.

\subsection{Synthetic data\label{sec:syntheticData}}

\ptitlenoskip{Data} We generated a dataset of 620 data points with two real-valued target attributes (attributes 1 and 2) and five binary descriptive attributes. We first sample 500 target values from the 2-D multivariate normal distribution $\cN (\bZero , \bI )$ and then embed three subgroups each consisting of 40 points into the data, see Fig.~\ref{fig:syntheticDataFigs}a. Each subgroup has distance 2 from the mean but a different covariance structure: the variance along the main eigenvector is much larger than the other. The first three descriptive attributes (attributes 3--5) contain the true labels for subgroups $p_1$ to $p_3$; the other two (attributes 6 and 7) take values randomly sampled from a Bernoulli distribution with $p = 0.5$.

\ptitle{Setup} We set the mean and covariance of the background model equal to the empirical values of the full data. First, we tested whether our method could retrieve the embedded patterns. We performed the two-step spread pattern mining process for three iterations, and at each iteration we selected the top pattern to update the background distribution. Second, we corrupted the descriptive attributes by randomly flipping every 0 and 1 with a certain probability. Then, we checked up to what noise level the subgroups can still be retrieved.

\begin{table}[t]
\centering
\caption{Change in SI for the top patterns over four iterations (\S \ref{sec:syntheticData}). All patterns have size 40. \label{table:siTop10In4Iter}}
\begin{tabular}{m{35mm}cccc}
Intention & SI Iter1 & Iter 2& Iter 3 & Iter 4\\ \hline
a3 = `1' & 48.35 & -1.13 & -1.13 & -1.13 \\
a5 = `1' & 47.49 & 47.49 & -1.13 & -1.13 \\
a4 = `1' & 39.49 & 39.49 & 39.49 & -1.13 \\
a4 = `0' $\wedge$ a3 = `1' & 36.26 & -0.85 & -0.85 & -0.85 \\
a5 = `0' $\wedge$ a3 = `1' & 36.26 & -0.85 & -0.85 & -0.85 \\
a3 = `0' $\wedge$ a5 = `1' & 35.62 & 35.62 & -0.85 & -0.85 \\
a4 = `0' $\wedge$ a5 = `1' & 35.62 & 35.62 & -0.85 & -0.85 \\
a3 = `0' $\wedge$ a4 = `1' & 29.62 & 29.62 & 29.62 & -0.85 \\
a5 = `0' $\wedge$ a4 = `1' & 29.62 & 29.62 & 29.62 & -0.85 \\
a5 = `0' $\wedge$ a4 = `0' $\wedge$ a3 = `1' & 29.01 & 29.01 & -0.68 & -0.68
\end{tabular}
\end{table}

\begin{figure}[tp]
  \includegraphics[width=\columnwidth]{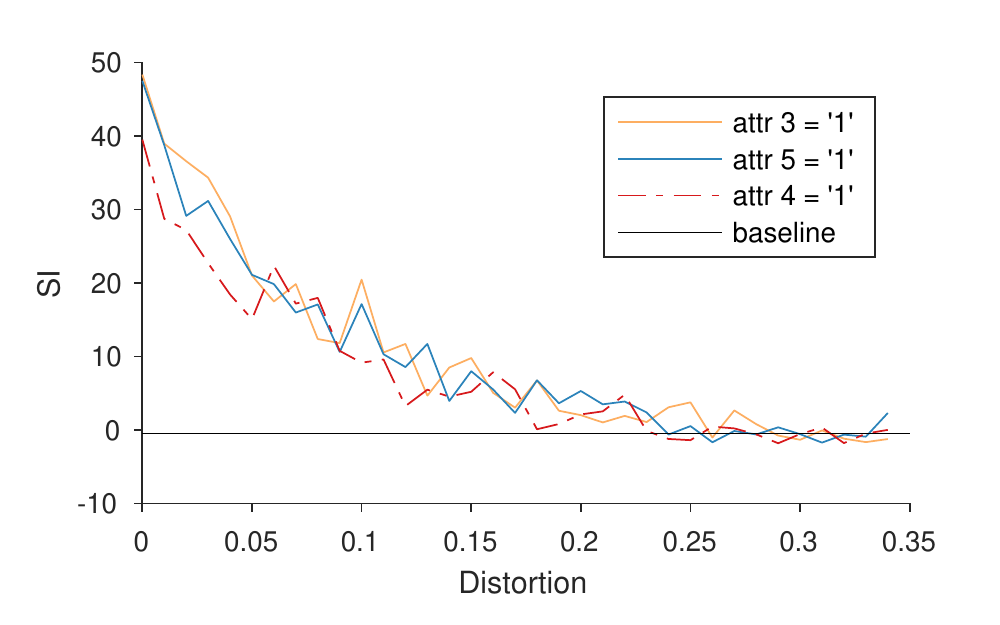}
  \caption{SI of subgroups in the synthetic data, (\S \ref{sec:syntheticData}), corresponding to true descriptions when adding and removing points randomly to the subgroups.\label{fig:robustness}}
\end{figure}

\ptitle{Results} Figures \ref{fig:syntheticDataFigs}b---\ref{fig:syntheticDataFigs}d show the top patterns in the first three iterations. Our method correctly found the embedded subgroups in the first three iterations by their displaced location from the expected center. It also retrieved the direction along which each subgroup's spread differs most from the full data covariance. Of course this is not so surprising, because for each embedded subgroup there is a description attribute setting the subgroup apart from the rest of the data.

To study the mining process in more detail, Table~\ref{table:siTop10In4Iter} shows the change in SI for the top 10 patterns from the first iteration in subsequent iterations. We observe that the three embedded subgroups were the highest-ranking patterns in the first three iterations (indeed they were the top 3 immediately because the subgroups induced by the true descriptions stand out so clearly from the rest of the data).

Once they were selected and used to update the background distribution, in the subsequent iterations the SI of the embedded subgroup patterns, and the SI of the derived patterns, dropped and remained low afterwards. Hence, updating the background distribution and the influence that should have on the IC scores of patterns worked as expected.

It can be observed also that the subgroups with more complex descriptions (e.g., a4 = `0' $\wedge$ a3 = `1') have lower SI, even though the extensions are equivalent to the corresponding $\text{a}_i$ = `1' pattern. This is because their DL is higher, while their extension is equivalent. Note that non-redundancy in the description is indeed achieved naturally in a principled manner. Also worth noting is that the SI can be negative. This is due to that the IC is based on a probability density and not a mass.

The result of the retrieval experiment with noise added to the description attributes is given in Fig.~\ref{fig:robustness}. We find that all embedded patterns can still be recovered when the flipping probability is up to 0.22, and partially retrieved up to 0.25. These values correspond to adding a random set of points that is roughly three and four times the size of the embedded pattern (e.g., $(1-0.25)\cdot40=30$ vs. $0.25\cdot480=120$). We conclude that the method is quite robust against noise.


\begin{figure*}[tp]
  \centering
  \includegraphics[width=0.9\textwidth]{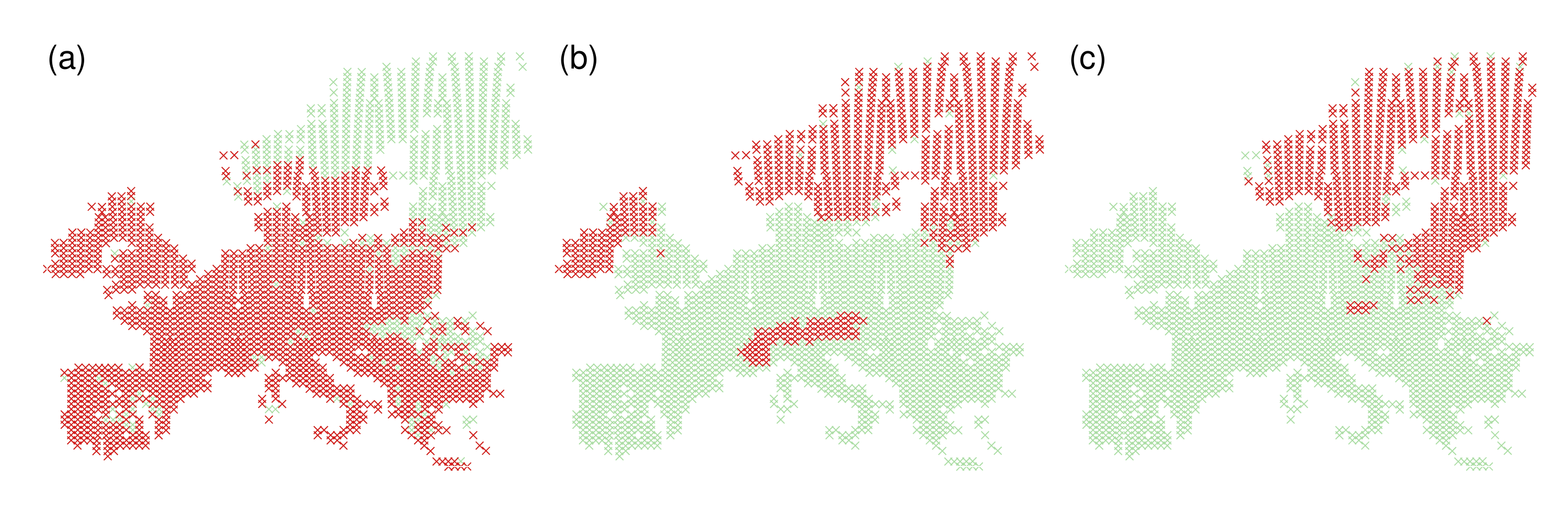}
  \caption{Explanation of what makes the first location pattern
  (Fig.~\ref{fig:topMammalPatterns}a) interesting (also see Fig.~\ref{fig:mammal1stPattern2}). Presence maps of the
  first three species in the full data: (a) wood mouse, (b) mountain hare, and (c) moose.\label{fig:mammal1stPattern1}}
\end{figure*}

\begin{figure}[tp]
  \includegraphics[width=\columnwidth]{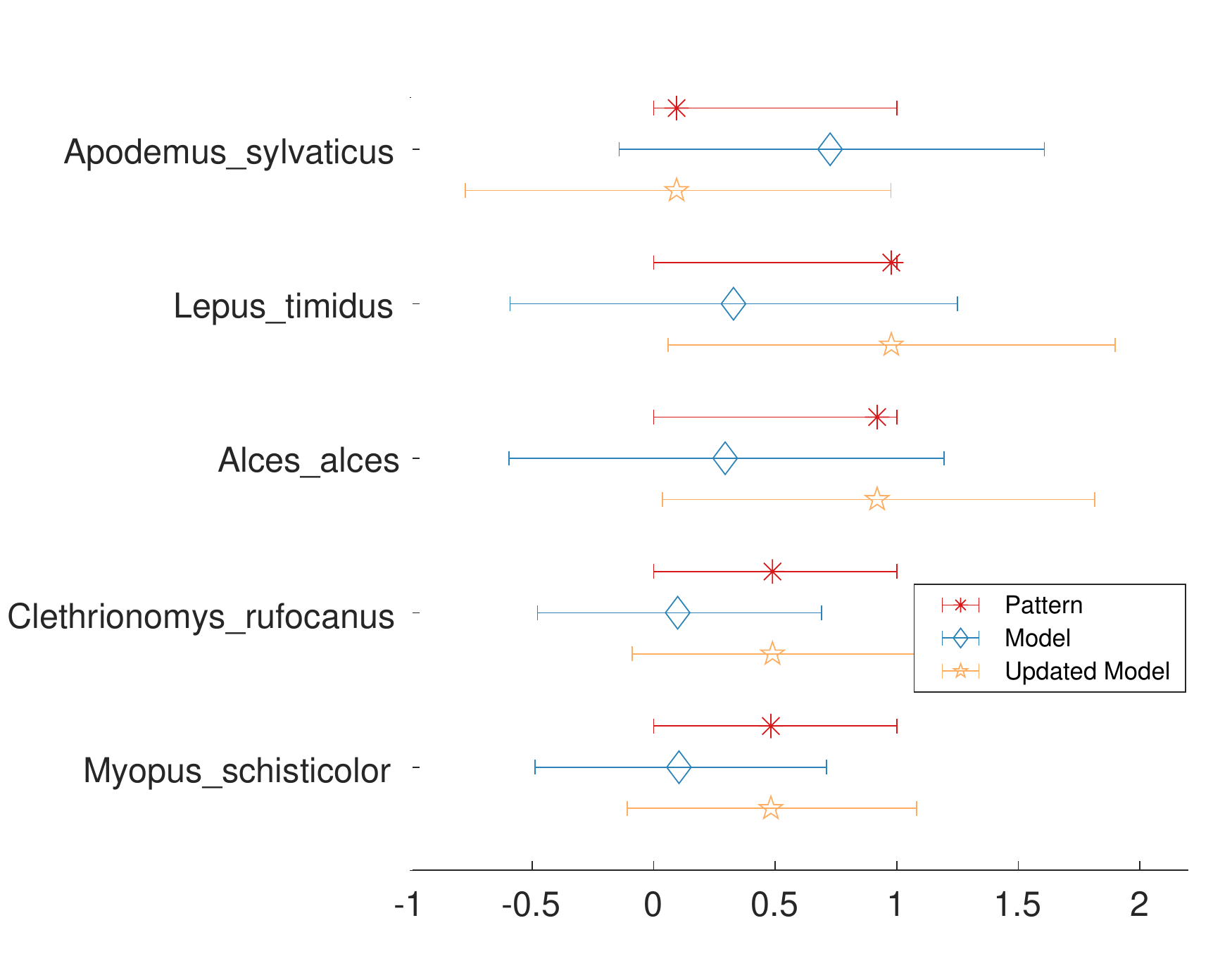}
  \caption{Explanation of what makes the first location pattern
  (Fig.~\ref{fig:topMammalPatterns}a) interesting (also see Fig.~\ref{fig:mammal1stPattern1}). Observed and expected
  mean and $95\%$ confidence interval of the most surprising species as ranked
  by SI.\label{fig:mammal1stPattern2}}
\end{figure}

\begin{figure*}[tp]
  \centering
  \includegraphics[width=0.9\textwidth]{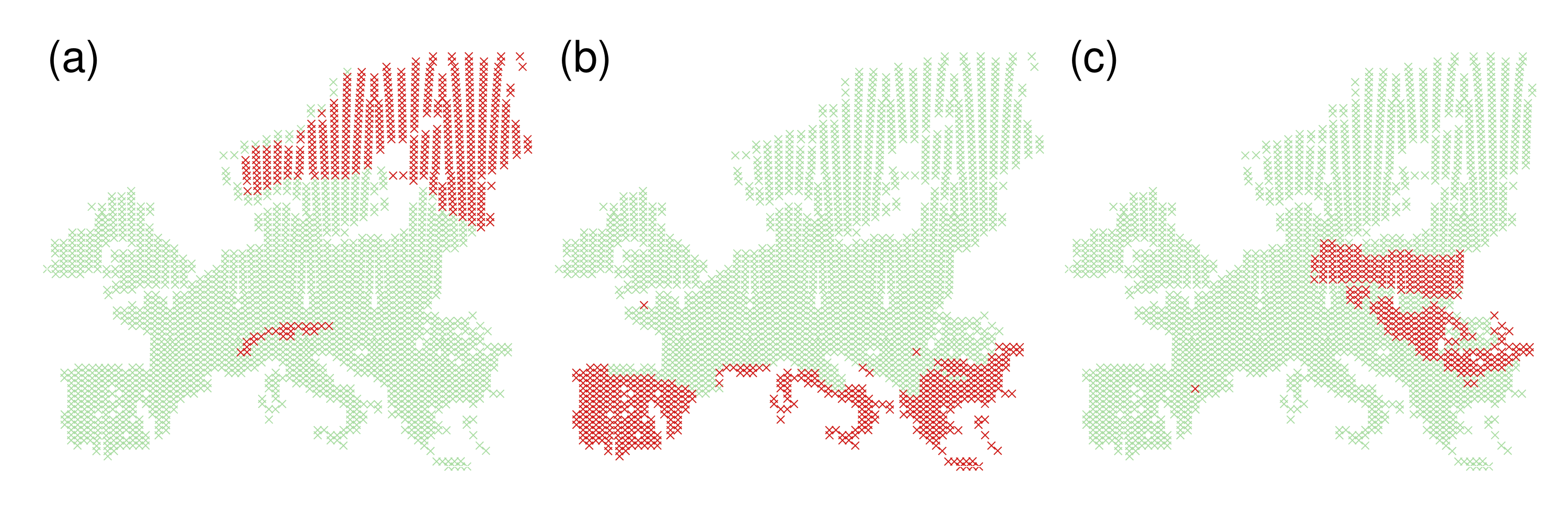}
  \caption{Extensions of the top three patterns found in the Mammal data (\S
  \ref{sec:mammalData}), (a) The first pattern covers northern Europe and part
  of the Alps. The intention is `mean temperature in March $\leq -1.68\ ^{\circ}$C'. (b) The second pattern covers the very south of Europe.
  Its intention is `average monthly rainfall in August $\leq 47.62$ mm'.
  (c) The third pattern covers parts of eastern Europe. The intention is
  `average monthly rainfall in October $\leq 45.25$ mm \emph{and}
  mean temperature of wettest quarter $\geq 16.32\ ^{\circ}$C'.
  \label{fig:topMammalPatterns}}
\end{figure*}

\subsection{Mammal data\label{sec:mammalData}}

\ptitlenoskip{Data} The mammal data encompasses data from The Atlas of European Mammals and from WorldClim.org, as preprocessed by Heikinheimo et al.\@ \cite{2007Heikinheimo}. It contains records about the presence of species in 2220 cells located on a grid that covers Europe. Each record contains the geolocation, binary labels for the presence/absence of 124 mammals, as well as 67 climate condition indicators.

\ptitle{Setup} We used the presence/absence indicators as target attributes and climate indicators for descriptions. The location information was used only for visualization and interpretation. We again set the initial mean and covariance parameters of the background model equal to the empirical values.

We found that for binary target attributes, spread patterns are not truly interesting. This makes sense, because the variance of a Bernoulli random variable is uniquely determined by the mean. Hence, a spread pattern becomes a one dimensional location pattern. That the attributes are binary is another form of background knowledge that could in principle be incorporated into the method, but it would lead to different derivations and we did not study this. Instead, we studied only location patterns on this data.

\ptitle{Results} The geographic locations of the data points part of the subgroup for the top patterns found in the first three iterations are visualized in Fig.~\ref{fig:topMammalPatterns}. The subrgoup intentions (combination of values that specifies the subrgoup) are given in the caption. The top pattern corresponds to locations that are relatively cold in late winter. In contrast, the second pattern covers locations that have an extremely dry summer, while the third pattern covers locations with a dry autumn and warm conditions in the months when most rain falls (which is the summer in that area).

We further investigated the distribution of the mammals within the subgroups. Fig.~\ref{fig:mammal1stPattern2} shows the mean values for the first pattern, and the mean and $95\%$ confidence interval for the background model for the top five mammal species ranked by SI. Figures~\ref{fig:mammal1stPattern1}a--c show the actual occurrences of the top three species across Europe. The species ranked first is the wood mouse, which is wide-spread in the middle and southern Europe but not in the northern areas.
 The second species is the mountain hare, whose habitat mostly coincides with the area associated to the found location pattern. This indicates it thrives under harsh temperature condition. The third species, moose, is also wide-spread mostly in the same area.
 
 By contrasting these ground-truth location maps for the species (Figs.~\ref{fig:mammal1stPattern1}a--c) against the subgroup location map (Fig.~\ref{fig:topMammalPatterns}), we find that indeed this pattern could be highly informative. However, while the description is concise, the displacement in the target space does not appear to be sparse (it covers many species). To comprehend the pattern in full, one should look at all the attributes where the mean deviates from the expectation, not just at the top five. This means fully understanding the pattern is somewhat difficult.

Finally, notice that these three species correlate and the background model already accounts for that. Hence, the IC of the subgroup is much less than the sum over the three attributes if they would be considered individually. Nonetheless, the IC is very high.

\begin{figure*}[tp]
  \centering
  \includegraphics[width=0.93\textwidth]{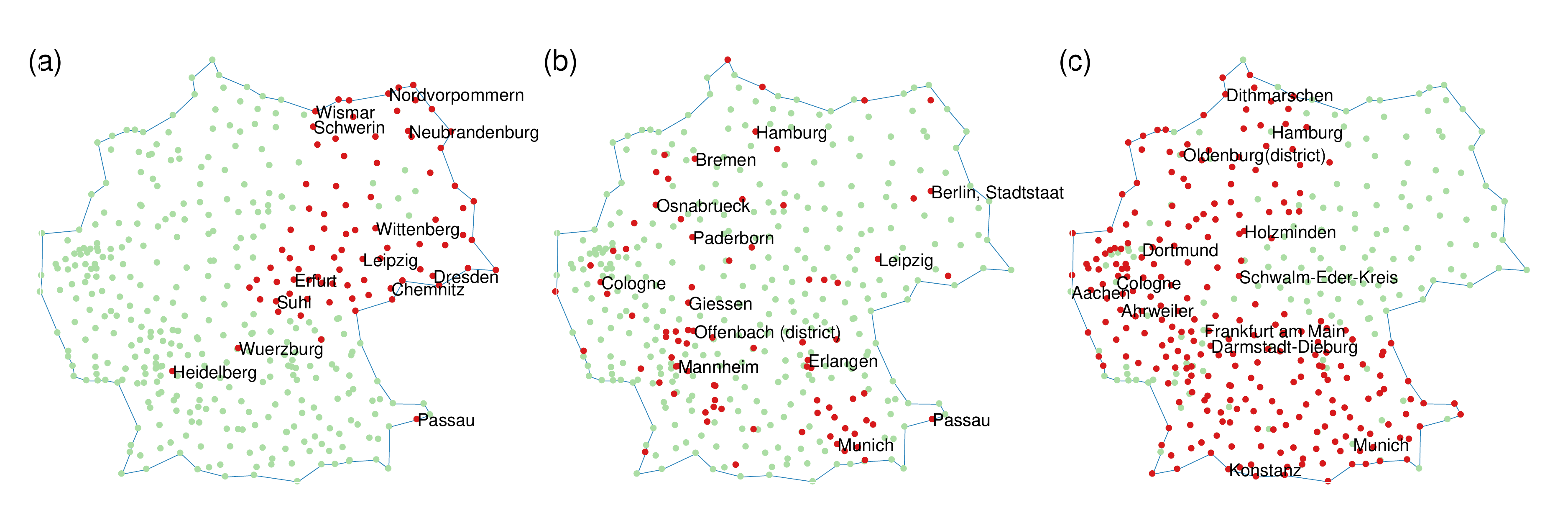}
  \caption{Geographic locations of data points covered by the top subgroup patterns found in the first three iterations on the Socio-economics data (\S \ref{sec:socioEconomicsData}): (a) ``Children Pop.\@ $<=$ 14.1'', (b)
``Middle-aged Pop.\@ $>=$ 26.9'', (c) ``Children Pop.\@ $>=$ 16.4''. The contents of these patterns is roughly as follows: (a) Low numbers of children are present in Eastern Germany, as well as in three cities with a very high percentage of students (Heidelberg, Passau, Wuerzburg). Here the Left party is popular is popular at the expense of all other parties (Fig.~\ref{fig:socioEcoPattern1Figs}a). (b) These are larger cities with relatively many jobs. Here the Green party is more popular at the expense of Left. (c) This subgroup is almost the complement of (a), but not quite (e.g., Saarland and smaller cities in the Ruhr area are not covered). Here Left is impopular and all others are more popular than the country-wide averages.
\label{fig:socioEcoPatterns}}
\end{figure*}

\begin{figure*}[tp]
  \centering
  \includegraphics[width=0.9\textwidth]{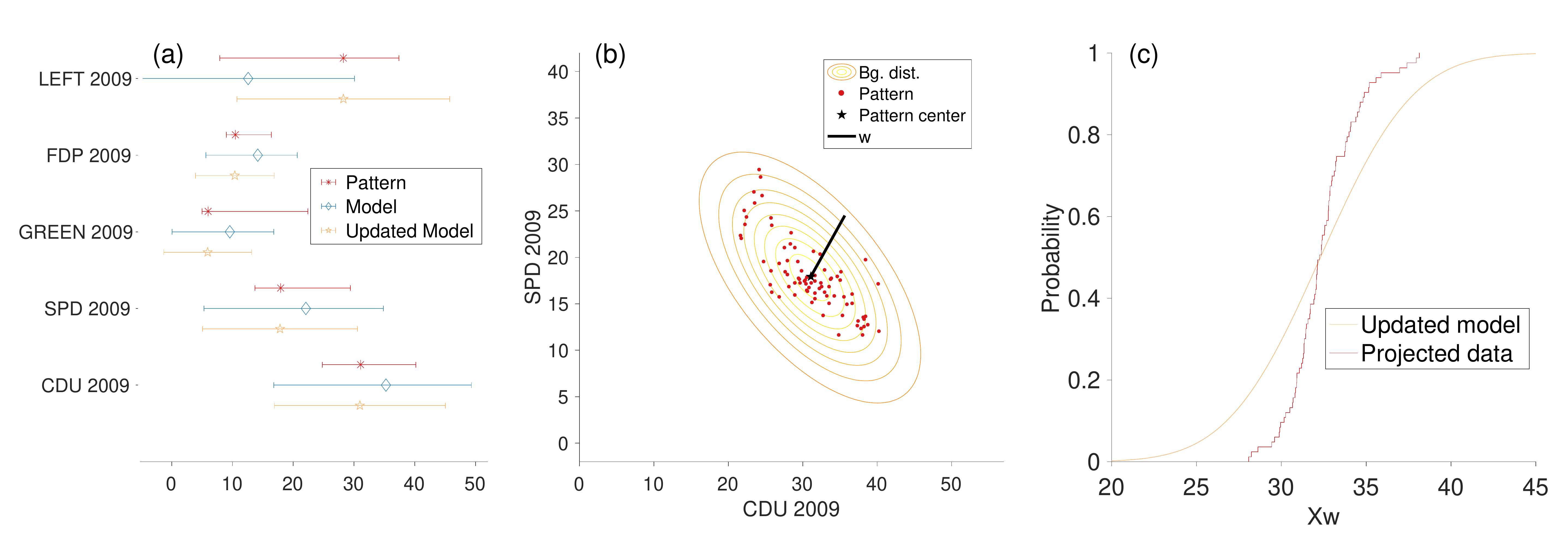}
  \caption{Spread pattern corresponding to the top pattern in Socio-economics data
(\S \ref{sec:socioEconomicsData}, Fig.~\ref{fig:socioEcoPatterns}a). (a)
Expected vs. observed distribution of the subgroup. the y-axis is ranked by SI,
from top to bottom. (b) Expected vs. observed distribution for the pair of
attributes with highest SI, after updating the background model with the
location pattern. The contour plot shows the found weight vector (black line)
along which the spread of the subgroup (red dots) has largest difference from
the background model (contour lines). (c) The marginal CDF of background
distribution and subgroup along $\bw$ after updating
location.\label{fig:socioEcoPattern1Figs}}
\end{figure*}

Although not shown, we repeated this exercise for the second and third pattern. The subgroup patterns appear to be informative. For example, ranked by SI, the most surprising species for the second pattern are the absence of the stoat and the bank vole,
who prefer a moist environment, and the presence of the Iberian hare,
who indeed lives exclusively in the area of the pattern. Thus, our method appears to find geographically meaningful location patterns that reveal the relationship between climate conditions and sets of animals that are absent/present in the corresponding area.

\subsection{Socio-economics data case study\label{sec:socioEconomicsData}}
\ptitlenoskip{Data} The German socio-economic dataset \cite{2013Boley} consists of socio-economic records of 412 administrative districts in Germany. The features are divided into three groups: election voting counts, age distribution, and workforce distribution.
The voting percentages of the five largest political parties (CDU/CSU, SPD, FDP, Greens, and Left) in the 2009 German elections are also included. We added the geographic coordinates of each district center ourselves.

\ptitle{Setup}  We used the vote count attributes as targets and the age and the work force attributes for the descriptions. Geolocations were used only for interpretation. Again, we set the initial mean and (co-)variance for the background distribution equal to the empirical values. In this case, that means we assume a user initially knows the overall voting behavior of the 2009 German elections.

We again performed three iterations of the subgroup discovery algorithm, but this time we studied both the location and the associated spread pattern in each iteration. To increase interpretability, we enforced a 2-sparsity constraint on $\bw$, by optimizing it for each pair of target attributes separately and then selecting the result with the highest SI.

\begin{figure*}[tp]
  \centering
  \includegraphics[width=\textwidth]{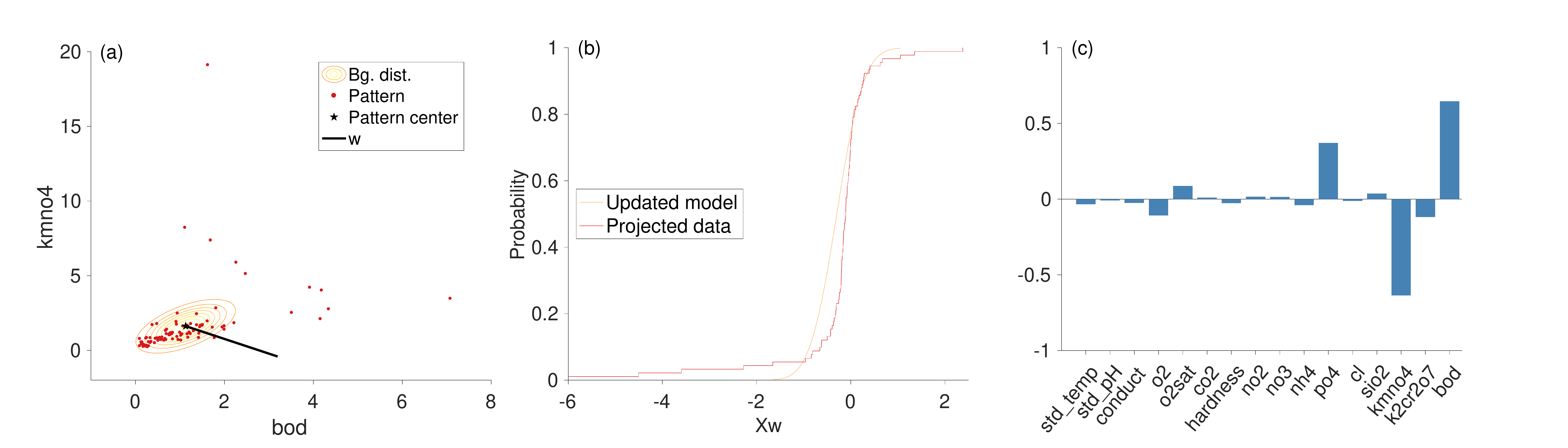}
  \caption{Top spread pattern found in the Water quality data (\S
  \ref{sec:waterQualityData}). (a) Subgroup vs. background
  distribution, along with the optimal projection vector $\bw$, projected on the
  two axes with highest weights. (b) CDF of subgroup and model along $\bw$. (c)
  The weight vector $\bw$ itself.\label{fig:waterQualityPattern2}}
\end{figure*}

\begin{figure}[tp]
  \centering
  \includegraphics[width=0.96\columnwidth]{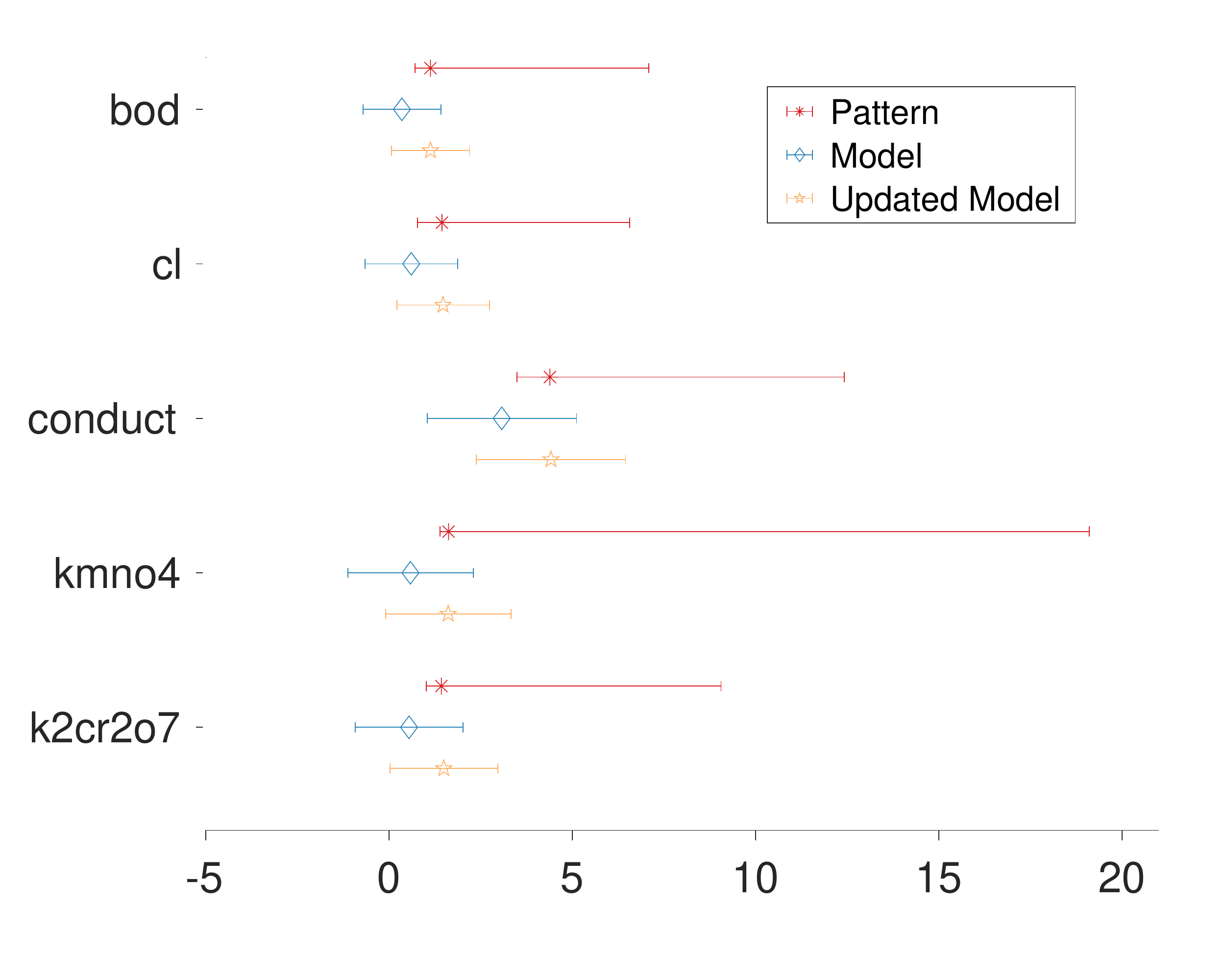}
  \caption{Observed and expected distribution of the top location pattern found
  in the Water quality data (\S \ref{sec:waterQualityData}), before and after
  updating location.\label{fig:waterQualityPattern1}}
\end{figure}

\ptitle{Results} Fig.~\ref{fig:socioEcoPatterns} shows the top location patterns found, and Fig.~\ref{fig:socioEcoPattern1Figs} some explanation and the spread pattern for the top location pattern. Comparing the distribution of the pattern against the expected distribution under the model (Fig.~\ref{fig:socioEcoPattern1Figs}a, red and blue lines), we observe that the voting behavior in the corresponding districts deviates substantially from the full population: more votes for Left, fewer for all others. The intention of the pattern corresponds to districts with relatively few children; from the map we see the extension covers mainly East Germany.

Once we update the background distribution with the location pattern, the model mean of the pattern becomes the observed mean, see Fig.~\ref{fig:socioEcoPattern1Figs}b. Given the updated background distribution, we find that the spread pattern with highest SI is related to the covariance between the social democrats (SPD) and Christian democrats (CDU), with weight vector $(0.5704, 0.8214)$ (see Fig.~\ref{fig:socioEcoPattern1Figs}c).\footnote{The 2d-contour plot of the subgroup is aggregated as the average pdf of the background model for each data point in the subgroup. The mean and covariance are the sub-vector and the sub-matrix that correspond to attributes indicated by the weight vector. This visualization is not fully accurate, as not all points have the same parameters. A single multivariate normal cannot represent the background model accurately.} As visualized in Fig.~\ref{fig:socioEcoPattern1Figs}d, the variance in this direction is much smaller than expected. Of course since the votes add up to a constant, under the model we also expect negative correlations between the parties, but for this subgroup the anti-correlation is much stronger than expected. This indicates these parties really appear to battle for the same voters. However, we are not sufficiently knowledgeable of German politics to judge whether this is a solid observation.

Fig.~\ref{fig:socioEcoPatterns} also shows the extensions for the top patterns in the second and third iterations. The second pattern has intention ``Middle-aged Pop.\@ $>=$ 26.9'' and contains large cities. Within those districts, the Green party has relatively high vote counts, which comes at the expense of the Left party. The third pattern, ``Children Pop.\@ $>=$ 16.4'', is mostly a complementary pattern to the first one (see Fig.~\ref{fig:socioEcoPatterns}a,c), except that many of the big cities (Munich, Berlin, Cologne, etc.) fall exactly between the two thresholds ($> 14.1, < 16.4$). The third pattern indeed covers locations where Left is unpopular and all other parties receive relatively many votes compared to the background model. In both the second and third location pattern, the corresponding spread pattern is a similar low-variance pattern as in Fig.~\ref{fig:socioEcoPattern1Figs}. In our subjective opinion, these patterns appear to convey potentially highly interesting insights into this data.

\subsection{Water quality data case study\label{sec:waterQualityData}}
\ptitlenoskip{Data} The River Water Quality dataset \cite{2000Dzeroski} consists of 1060 water quality records sampled from rivers in Slovenia.
Each record contains measured values for 16 physical/chemical parameters and 14 bioindicators (7 plants, 7 animals), including a list of all taxa present and their density. The density of each taxon is recorded by an expert biologist at three different qualitative levels,
where 1 means the taxon occurs incidentally, 3 frequently, and 5 abundantly.

\ptitle{Setup} We use the 16 physical/chemical parameters as targets and the 14 bioindicators as descriptors. Mean and \mbox{(co-)}variance of the initial background distribution were set to the empirical values.

\ptitle{Results}
The top location pattern has intention ``Amphipoda Gammarus fossarum $<=$ 0 AND Oligochaeta Tubifex  $>=$ 3'' and covers 91 records. Fig.~\ref{fig:waterQualityPattern1}a shows that the water samples fulfilling the description have an above-average biological oxygen demand (BOD), chlorine concentration (Cl), electrical conductivity, as well as K$_2$Cr$_2$O$_7$ and KMnO$_4$ (indicating chemical oxygen demand, COD).

In the second step our method finds, without enforcing it, a sparse weight vector placing high weights on BOD and KMnO$_4$ (Fig.~\ref{fig:waterQualityPattern1}d). The contour plot (Fig.~\ref{fig:waterQualityPattern1}b) indicates that along the most interesting spread direction, $\bw$, the variance of the subgroup is much larger than expected. The CDF in Fig.~\ref{fig:waterQualityPattern1}c also confirms this. The main conclusion here is that, although the identified patterns are typically subgroups that are displaced from the center of the data, which is typically associated with having a smaller variance in comparison to the full data, it is also possible to find spread patterns corresponding to surprising higher-variance directions.

\subsection{Scalability}

We have not analyzed the algorithmic complexity of mining optimal location and spread patterns in detail, nor have we studied extensively how to find good solutions in practice. The computation time of the beam search algorithm can be controlled through the search parameters (number of solutions kept at each iteration, discretization strategy for numerical attributes, maximum number of conditions for the description) and it employs a timer. Of course it may not find the optimal pattern, but this strategy allows it to work on data of any size and dimensionality. Likewise, the heuristic solution to mine spread patterns typically outputs a pattern in very little time.

Notice that for both algorithms, the runtime is linear in the number of data points (i.e., to do the exact same computations on larger data is linear). One may add attributes without affected the computation time at the mining stage (background model discussed below), but of course to include them in candidate descriptions leads to an exponential growth in number of possible subgroup definitions. We feel it would be pointless to include a runtime experiment for these steps, as it is not feasible to compute the optimal solutions as a comparison, except on very small data.

What we can analyze is the runtime of fitting the background distribution. For all four real-world datasets, we mined location and spread patterns and measured the time it took to find the new MaxEnt distribution incorporating both previous and the newly identified pattern, for 20 iterations. The results are presented in \ref{table:scalability}. We find that after insertion of 10--20 location patterns, the time it takes to find the MaxEnt distribution becomes noticeable. This may not be so surprising, as there are at least $\dy$ new constraints every time we insert a new location pattern. For the Mammals data, which has target dimension 124, the time quickly grows to durations that cannot be considered acceptable for interactive use. We also observe that for spread patterns, this problem does not occur because they are by definition of low rank (the weigth vector is not necessarily sparse but it is only a one-dimensional projection).

\begin{table}[t]
\centering
\caption{Runtime measurements to update the background distribution with identified patterns. First row shows time (in seconds) to fit the initial distribution, consecutive rows the time until convergence when incorporating additional patterns. As the updates for location and spread patterns are  different, these are reported independently (columns 2--5 and 6--9). Data sets: German Socio-Economics (GSE; $n = 412$, $\dx = 13$, $\dy = 5$), Water Quality (WQ; $n = 1060$, $\dx = 14$, $\dy = 16$), Crime (Cr; $n = 1994$, $\dx = 122$, $\dy = 1$), Mammals (Ma; $n = 2220$, $\dx = 67$, $\dy = 124$). \label{table:scalability}}
\begin{tabular}{lccccccc}
 & \multicolumn{4}{c}{Location pattern} & \multicolumn{3}{c}{Spread pattern} \\ 
Iteration & \multicolumn{1}{|c}{GSE} & WQ & Cr & Ma & \multicolumn{1}{|c}{GSE} & WQ & Cr \\ \hline
Init & 9.167 & 8.640 & 9.714 & 8.453 \\
1 & 0.13 & 0.16 & 0.12 & 13.72 & 0.10 & 0.10 & 0.11\\ 
2 & 0.09 & 0.16 & 0.08 & 33.09 & 0.08 & 0.05 & 0.08\\ 
3 & 0.12 & 0.31 & 0.09 & 62.61 & 0.06 & 0.12 & 0.09\\ 
4 & 0.25 & 0.52 & 0.11 & 120.44 & 0.11 & 0.13 & 0.13\\ 
5 & 0.33 & 0.92 & 0.16 & 184.33 & 0.14 & 0.18 & 0.20\\ 
6 & 0.49 & 1.41 & 0.19 & 250.23 & 0.19 & 0.19 & 0.27\\ 
7 & 0.68 & 1.94 & 0.30 & 399.90 & 0.26 & 0.32 & 0.44\\ 
8 & 0.91 & 2.57 & 0.41 & 602.54 & 0.37 & 0.36 & 0.50\\ 
9 & 1.16 & 3.07 & 0.56 & 796.38 & 0.38 & 0.37 & 0.65\\ 
10 & 1.49 & 4.00 & 0.80 & 1130.81 & 0.42 & 0.46 & 0.83\\ 
11 & 1.69 & 5.05 & 1.02 & - & 0.42 & 0.49 & 1.07\\ 
12 & 1.95 & 6.17 & 1.23 & - & 0.52 & 0.57 & 1.32\\ 
13 & 2.56 & 7.48 & 1.52 & - & 0.63 & 0.65 & 1.62\\ 
14 & 2.76 & 9.04 & 1.95 & - & 0.68 & 1.16 & 2.09\\ 
15 & 3.17 & 10.60 & 2.60 & - & 0.72 & 1.00 & 2.86\\ 
16 & 3.51 & 11.92 & 3.41 & - & 0.81 & 1.06 & 3.42\\ 
17 & 4.40 & 14.06 & 4.15 & - & 1.12 & 1.38 & 5.01\\ 
18 & 4.94 & 15.95 & 5.34 & - & 1.17 & 1.47 & 5.69\\ 
19 & 4.99 & 17.92 & 6.66 & - & 1.07 & 1.57 & 6.30\\ 
20 & 5.58 & 19.97 & 6.71 & - & 1.24 & 1.92 & 6.65
\end{tabular}
\end{table}

\section{Related Work}\label{sec:relatedwork}

The pattern syntax introduced in this paper can be considered a type of \emph{Exceptional Model Mining} (EMM)
\cite{2016Duivesteijn,2008Leman}. EMM can be seen as a multi-target generalization of \emph{Subgroup Discovery} (SD) \cite{1996Kloesgen},
which is a single-target supervised form of \emph{Pattern Mining} \cite{2005Morik}: the broad subfield of data mining where only a part of the data is described at a time, ignoring the coherence of the remainder.

Tasks similar to SD are
Contrast Set Mining \cite{2001Bay} and Emerging Pattern Mining
\cite{1999Dong}.  Both these tasks have not been considered for multiple target
attributes simultaneously, and hence differ from the current paper in that they do not directly help in understanding interactions between variables. The relationships between Contrast Set Mining, Emerging Pattern Mining, and SD are extensively described in
\cite{2009KraljNovak}.

Distribution Rules \cite{2006Jorge} can be seen as an early
instance of EMM with only one target.
Umek et al.\@ \cite{2011Umek} do consider SD
with multiple targets.  They approach the attribute partition
in the reverse way of EMM: candidate subgroups are generated by
agglomerative clustering on the targets, and predictive modeling on the
descriptors strives to find matching descriptions.

\emph{Redescription Mining} by Galbrun et al. \cite{2012Galbrun} is the closest related work to this paper. It considers the case where a dataset contains two
distinct parts, describing the same entities from two different viewpoints.
Redescription Mining treats these two parts symmetrically: it seeks descriptions
inducing the same subgroup, resulting in a rule of the form $A\simeq B$.  In contrast, we consider the setting where the two parts play distinct roles: one part contains description attributes on which subgroups are defined, the other part forms the numeric data which we aim to learn about and hence on which the informativeness of subgroups is evaluated. This then results in rules of the form $A\Rightarrow B$.

Interestingly, Galbrun et al. \cite[Fig.\@ 8, Tab.\@ 6, 7]{2012Galbrun} also considered the problem of `biological niche finding' on the Mammal data. However, none of the subgroups they report are the same as ours. Their version of the data also encompasses a slightly larger region, but it is anyway unsurprising that results are quite different. The score function in Redescription Mining is not based on how much the subgroups stand out from the overall data, but only on the accuracy of the redescription and its cover. Hence, we did not further compare the results of our method with theirs.

`\emph{Subjective Interestingness}' was first used in the context of Association Rule Mining \cite{PaT:98,silberschatz1996}.
These papers formalized the prior belief of a user in a belief system, and sought
association rules that contrasted with these beliefs.
We base our approach on the more recent and systematic approach named FORSIED \cite{debie2011,debie2013}. This framework has been applied successfully to a variety of data mining problems, such as mining relational patterns \cite{lijffijt2016}, community detection \cite{vanleeuwen2016}, clustering \cite{kontonasios2015}, and dimensionality reduction \cite{kang2016a}. Maximum Entropy modeling for real-valued data has also been studied before \cite{KVD:11}, in order to compute the significance of the Weighted Relative Accuracy in SD. That method targets a different pattern syntax  than what is introduced here and does not apply to EMM.

Finally, Boley et al. \cite{boley2017} recently introduced a score function for single-target SD where a reduction in variance adds to the interestingness score of a subgroup. While their approach is less general and the interestingsness score arguably less principled, they do study the algorithmic complexity of the problem in detail and derive a tight-optimistic-estimator-based branch and bound algorithm to find the globally best subgroup pattern very efficiently.

\section{Discussion and Conclusion}\label{sec:discussion}

Numerous unsupervised methods exist to make sense of real-valued datasets,
most notably methods for dimensionality reduction and clustering.
Labels (or more generally description attributes as in this paper) associated with the data points are then often used to interpret these results,
e.g., by measuring enrichment of certain labels within a cluster,
or by coloring data points in a scatter plot of a 2-D
projection of the data with a color depending on the labels of the points,
for subsequent visual inspection.
However, whether such analyses provide explanations or insights is a matter of coincidence:
there is no a priori reason that clusters should be enriched,
and there is no guarantee that equally colored points are grouped in a scatter plot.

Here, we propose an alternative approach,
in directly using the description attributes to guide the search for surprising multivariate relations in the data. Resulting subgroups are then automatically explained well by the descriptions.
Our approach contrasts with traditional supervised methods in focusing on \emph{local} patterns:
properties of the target attributes that apply only to subsets of the data defined in terms of conditions on their metadata.
Arguably, with increasing amounts and resulting inhomogeneity of datasets,
the importance of local patterns is bound to increase.

Our approach generalizes the literature on Subgroup Discovery and Exceptional Model Mining in being applicable for real-valued target attributes of arbitrary dimensionality, and in searching for multivariate local patterns across all these dimensions, including unusual covariance structures of subgroups in the data. Moreover, the interestingness of the patterns of this type is formalized in a rigorous manner, quantifying the amount of information the user gains by observing them. We have demonstrated that the resulting algorithms are effective and efficient, in theory and in practice.

In further work, we plan to remove the dependency on third party tools (Matlab and Cortana) and produce a standalone version of the method for public dissemination. Furthermore, it would be interesting to study similar pattern syntaxes for binary, categorical, and mixed sets of target attributes. Besides, although we have little hope to improve the search for optimal spread patterns, it may be feasible to devise a branch-and-bound approach to mine optimal location patterns efficiently. Indeed this appears to be the most relevant question to be addressed in the future. Finally, we aim to integrate this method with SIDE \cite{kang2016b,puolamaki2016}, our online tool for exploration of numerical data, which currently does not use any labels or description attributes.

\vspace{2mm}\noindent{\em Acknowledgements.} This work has been supported by the ERC under the EU's Seventh Framework Programme (FP/2007-2013) / ERC Grant Agreement no. 615517, FWO (project no. G091017N, G0F9816N), the EU's Horizon 2020 research and innovation programme and the FWO under the Marie Sk{\l }odowska-Curie Grant Agreement no. 665501, the Academy of Finland (decision 288814), and Tekes (Revolution of Knowledge Work project).

\bibliographystyle{IEEEtran}
\balance
\bibliography{paper}

\end{document}